\documentclass[journal]{IEEEtran}
\ifCLASSINFOpdf
\else
\fi
\usepackage{amscd}

\usepackage{amssymb}
\usepackage{amsmath}

\usepackage{latexsym}

\usepackage{textcomp}

\usepackage{mathabx}

\usepackage{multirow}% http://ctan.org/pkg/multirow
\usepackage{hhline}% http://ctan.org/pkg/hhline

\usepackage{epstopdf}
\usepackage{epsfig}
\usepackage{enumitem}
\usepackage[numbers,sort&compress]{natbib}

\usepackage{ifpdf}

\usepackage{amscd}

\usepackage{amssymb}
\usepackage{amsmath}

\usepackage{latexsym}

\usepackage{textcomp}

\usepackage{mathabx}

\usepackage{multirow}% http://ctan.org/pkg/multirow
\usepackage{hhline}% http://ctan.org/pkg/hhline

\usepackage{subfig}

\usepackage{threeparttable}

\usepackage{algorithm}

\usepackage{algorithmicx}

\usepackage{algpseudocode}

\usepackage{url}

\usepackage{color}

\begin{document}
%\title{Compressive Imaging History Identification}
\title{Forensic Discrimination between Traditional and Compressive Imaging Systems}

\author{Ali Taimori, Farokh Marvasti,~\IEEEmembership{Senior Member, IEEE}% <-this % stops a space

\thanks{A. Taimori and Farokh Marvasti are with the Electrical Engineering Department, Sharif University of Technology, Tehran 14588-89694, Iran (e-mails: alitaimori@yahoo.com; marvasti@sharif.edu).
}}% <-this % stops a space
\markboth{IEEE TRANSACTIONS ON INFORMATION FORENSICS AND SECURITY,~Vol.~x, No.~x, August~xxxx}%
{Shell \MakeLowercase{\textit{et al.}}: Bare Demo of IEEEtran.cls for IEEE Journals}



\maketitle

\begin{abstract}
Compressive sensing is a new technology for modern computational imaging systems. In comparison to widespread conventional image sensing, the compressive imaging paradigm requires specific forensic analysis techniques and tools. In this regards, one of basic scenarios in image forensics is to distinguish traditionally sensed images from sophisticated compressively sensed ones. To do this, we first mathematically and systematically model the imaging system based on compressive sensing technology. Afterwards, a simplified version of the whole model is presented, which is appropriate for forensic investigation applications. We estimate the nonlinear system of compressive sensing with a linear model. Then, we model the imaging pipeline as an inverse problem and demonstrate that different imagers have discriminative degradation kernels. Hence, blur kernels of various imaging systems have utilized as footprints for discriminating image acquisition sources. In order to accomplish the identification cycle, we have utilized the state-of-the-art Convolutional Neural Network (CNN) and Support Vector Machine (SVM) approaches to learn a classification system from estimated blur kernels. Numerical experiments show promising identification results. Simulation codes are available for research and development purposes.
\end{abstract}


% \fi
\IEEEpeerreviewmaketitle

\section{Introduction}
\IEEEPARstart{T}{he} emerging technology of Compressive Sensing (CS) has lead a new generation of imaging modalities called compressive imaging \cite{romberg2008imaging}. This kind of imaging provides attractive advantages in comparison to conventional models such as considerable reduction in hardware components, e.g. sensors and Analog/Digital (A/D) converters, reducing power consumption, and fast image acquisition time. These gains obtain due to the fact that conventional imagers operate generally at the sampling rate of Shannon-Nyquist, whereas compressive imaging systems work at a rate less than the available traditional ones, where computation play a key role.

The superiority of compressive sensing technology excites academic researchers and industry engineers to migrate from the current traditional imaging systems to compressively sensed-based ones. The Rice single pixel camera~\cite{duarte2008single} and its variants~\cite{WinNT}, and compressed sensing-based Magnetic Resonance Imaging (MRI) are paradigms of such attempts. In a compressive camera, sampling and compression processes are done simultaneously in one step~\cite{taimori2018adaptive}. In other words, compression directly performs at sensor level. Whereas, conventional imagers, at first, acquires an image at Shannon-Nyquist rate and then, if needed for storage/transmission purposes, discard redundant data by compression.

Due to advancement and deployment of compressive imaging systems in applications such as consumer electronics, environmental surveillance, remote sensing, and medicine in the near future \cite{WinNT}, the field of multimedia forensics require novel and well-tailored forensic techniques and tools to constantly answer questions arose from the whole society.

%\subsection{Considered Scenario}
\subsection{Considered Scenario and Related Work}
One of basic scenarios in image forensics is to distinguish traditionally sensed images from sophisticated compressively sensed ones \cite{chu2012forensic, chu2015compressive}. In this scenario, assume an image with an uncompressed file format such as Bitmap (BMP) be available for forensic analysis. Now, the question is that whether the image under investigation has been captured from a conventional camera or a compressively sensed-based imaging system. It is important to note that if the digital photo has been acquired from a conventional imaging system, it may has been compressed via a source coder such as the well-known JPEG compression or acquired in a raw manner without any compression. Therefore, in the considered scenario, three types of imaging systems are introduced, i.e. compressive imaging, conventional raw imaging without compression, and conventional imaging plus compression. Forensic identification of these imaging systems reveal the history of image acquisition such as the source camera that has been generated the questionable image and settings under which the image has been captured \cite{chu2015compressive}. From a forensic analyzer viewpoint, a tool for discriminating the above imaging systems helps to detect
\begin{list}{\labelitemi}{\leftmargin=0.5em}
\item traces about the source device which has acquired the image under investigation and
\item forged images which has been generated by a composition of both traditionally and compressively sensed images.
\end{list}
%\begin{itemize}
%  \item traces about the source device which has acquired the image under investigation,
%  \item forged images which has been generated by a composition of both traditionally and compressively sensed images.
%\end{itemize}

To date, information forensic science has experienced valuable studies about traditional imaging forensics both in theory and practice~\cite{farid2016photo, kavrestad2017guide, stamm2013information, piva2013overview}. Forensic analyses of conventional imaging under different scenarios can be categorized into two main groups. The first one includes approaches that identify the source has been captured an questionable images by assigning the image to a device make and model~\cite{lukas2006digital, tuama2016camera}. The second group tries to detect or localize forgery traces in digital images~\cite{taimori2015quantization, bahrami2015blurred}. In these approaches, discovering intrinsic signatures left from recorded images plays a key role for forensic investigators.

In spite of a relative maturity in conventional imaging forensics, compressive imaging forensics is in its infancy. To the best of our knowledge, the studies Chu et al. in~\cite{chu2012forensic, chu2015compressive} are the primary researches toward forensic analysis of compressive sensing. They discovered the empirical probability mass function of wavelet coefficients for different imaging systems have Laplace-like distributions with different location and diversity parameters. In their method, a 2-step thresholding-based decision making process is employed to differentiate between traditional sensing from compressive ones. The first step contains a distribution detector based on Maximum Likelihood (ML) estimator to discriminate raw images from JPEG2000 (JP2) and compressively sensed images. The second distribution-based detector gains Expectation Maximization (EM) algorithm to classify traditionally sensed JPEG2000 images from compressively sensed ones. It is important to note that the well-known JPEG2000 and most of compressive sensing algorithms utilize wavelet transform for coding \cite{sayood2017introduction, sadeghigol2016model}. This refers to the point that the confusion of these imaging systems is more probable. Therefore, the focus of Chu at al. method is on JPEG2000 standard among other image compression techniques.

Although the pioneer research~\cite{chu2015compressive} has its own importance, the filed of information forensics requires more analytical investigations. This issue is important especially for the emerging technology of compressive sensing, where forensic examiners require to identify different aspects of such an imaging model. Therefore, in the context of media forensics, we need to develop specific theoretical and applied studies for compressive imaging forensics.

\IEEEpubidadjcol

\subsection{Motivations and Contributions}
As mentioned above, compressive sensing modeling is of importance for forensic investigators. Such modeling helps for better understanding of the complex nature of compressive imaging process in both encoder and decoder sides. In this paper, we first mathematically and systematically model the imaging systems based on compressive sensing technology. Afterwards, a simplified version of the whole model is presented which is appropriate for forensic investigation purposes. We estimate the sophisticated nonlinear system of compressive sensing with a linear model. Based on the best knowledge of authors, this is the first modeling of compressive imaging.

Then, we model the imaging pipeline as a degradation system and mathematically show that different imagers have discriminative degradation blur kernels. Therefore, blur kernels of various imaging systems can be utilized as footprints for discriminating their sources. Various blind deconvolution approaches exist to estimate an unknown burring kernel as well as a sharp image version of the blurred image. To this intent, we blindly estimate blurring kernels for a database of different images captured from the three considered imagers via the deconvolution algorithm of \cite{levin2011efficient}. At the end, extracted features are fed to a supervised learning mechanism to train a identification system for discrimination. Briefly, the main contributions of this paper are
\begin{list}{\labelitemi}{\leftmargin=0.5em}
\item modeling of a general compressive imaging system,
\item deriving a simplified compressive camera model from the general model for forensic analysis purposes,
\item providing simple models for traditional image sensing,
\item revealing traces left by conventional/compressive imaging systems and demonstrating their difference, and
\item proposing an learning-based identification technique for discriminating conventional cameras from compressive ones based on the estimated characteristics.
\end{list}
%\begin{itemize}
%  \item modeling of a general compressive imaging system,
%  \item deriving a simplified compressive camera model from the general model for forensic analysis purposes,
%  \item providing simple models for traditional image sensing,
%  \item revealing traces left by conventional/compressive imaging systems and demonstrating their difference, and
%  \item proposing an identification technique for discriminating conventional cameras from compressive ones based on the discriminative characteristics.
%\end{itemize}

\subsection{Paper Organization}
This paper is organized as follows. In Section \ref{ProblemModelingAndFormulation}, we introduce three models for imaging, one for compressive sensing of images and other two models for conventional imaging encompassing raw imaging and conventional raw imaging plus JPEG2000 compression. A learning-based approach is suggested to forensically distinguish these types of imaging mechanisms in Section \ref{IdentificationMethodology}. In Section \ref{NumericalExperiments}, the proposed identification method is supported by numerical experiments. Finally, we conclude the paper in Section \ref{ConclusionsAndFutureResearches}.

\section{Problem Modeling and Formulation}
\label{ProblemModelingAndFormulation}
\begin{figure}[!t]
\centering
\includegraphics[width=\linewidth]{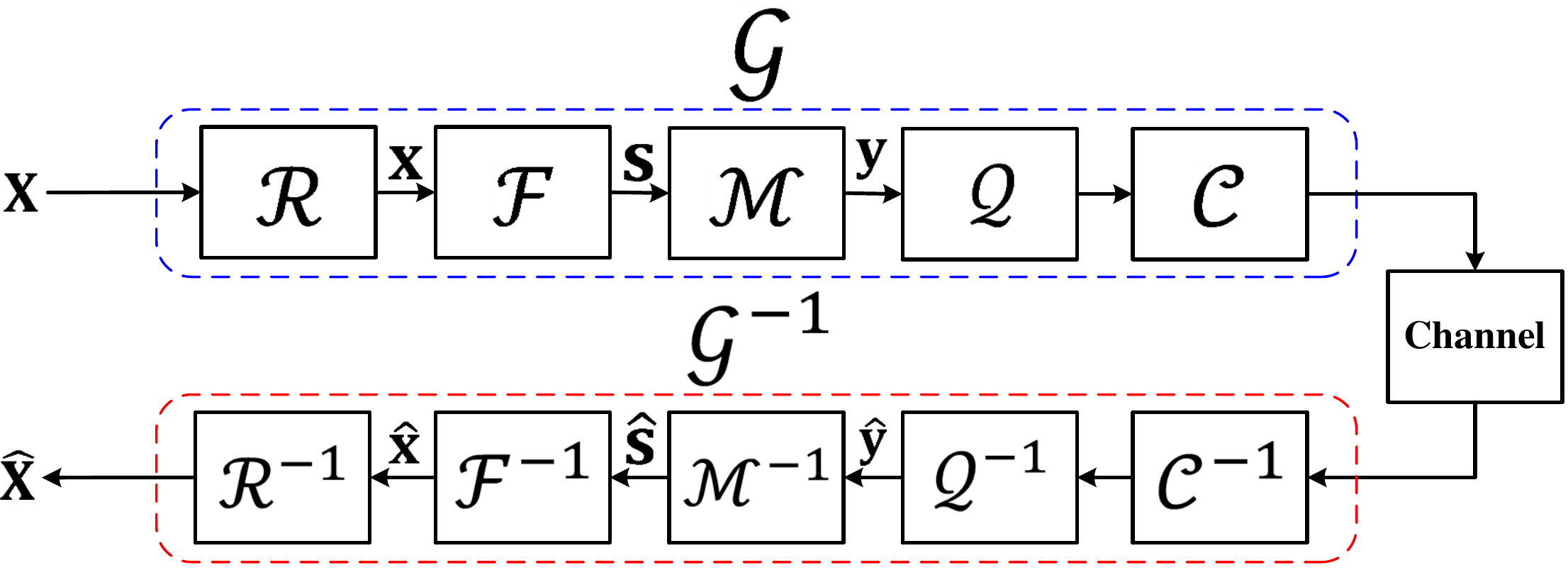}
\caption{Modeling of the encoder $\mathcal{G}$ and the decoder $\mathcal{G}^{-1}$ in a compressive imaging system using some linear and nonlinear mathematical operators.}
\label{CompressiveImagingModel}
\end{figure}

\subsection{Compressive Imaging Model}
\begin{figure*}[!t]
\centering
%%----start of first subfigure----
\subfloat[]{
\label{fig:subfig:a} %% label for first subfigure
\includegraphics[width=5cm]{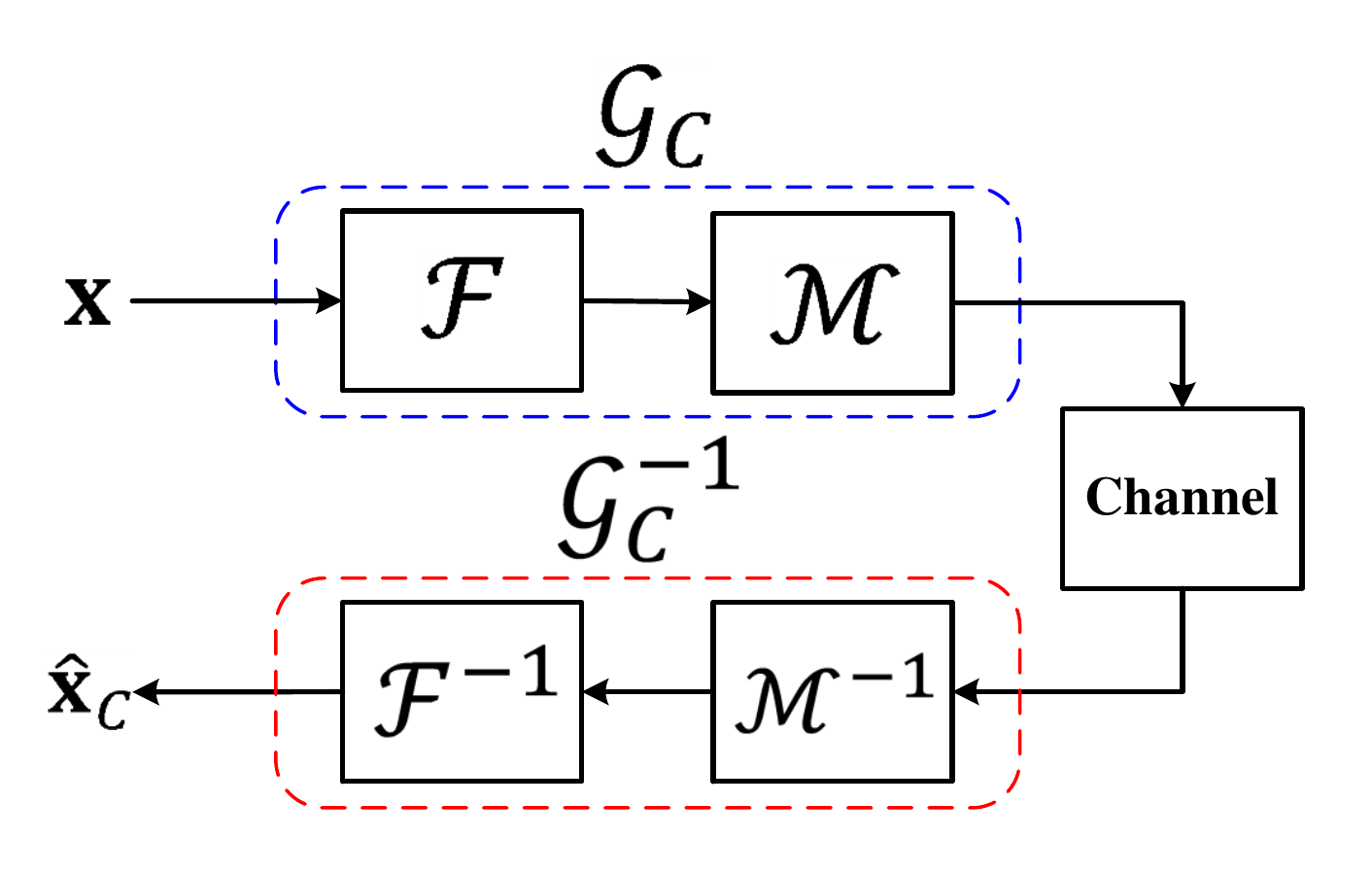}}
\hspace{0.5cm}%
%%----start of second subfigure----
\subfloat[]{
\label{fig:subfig:b} %% label for second subfigure
\includegraphics[width=5cm]{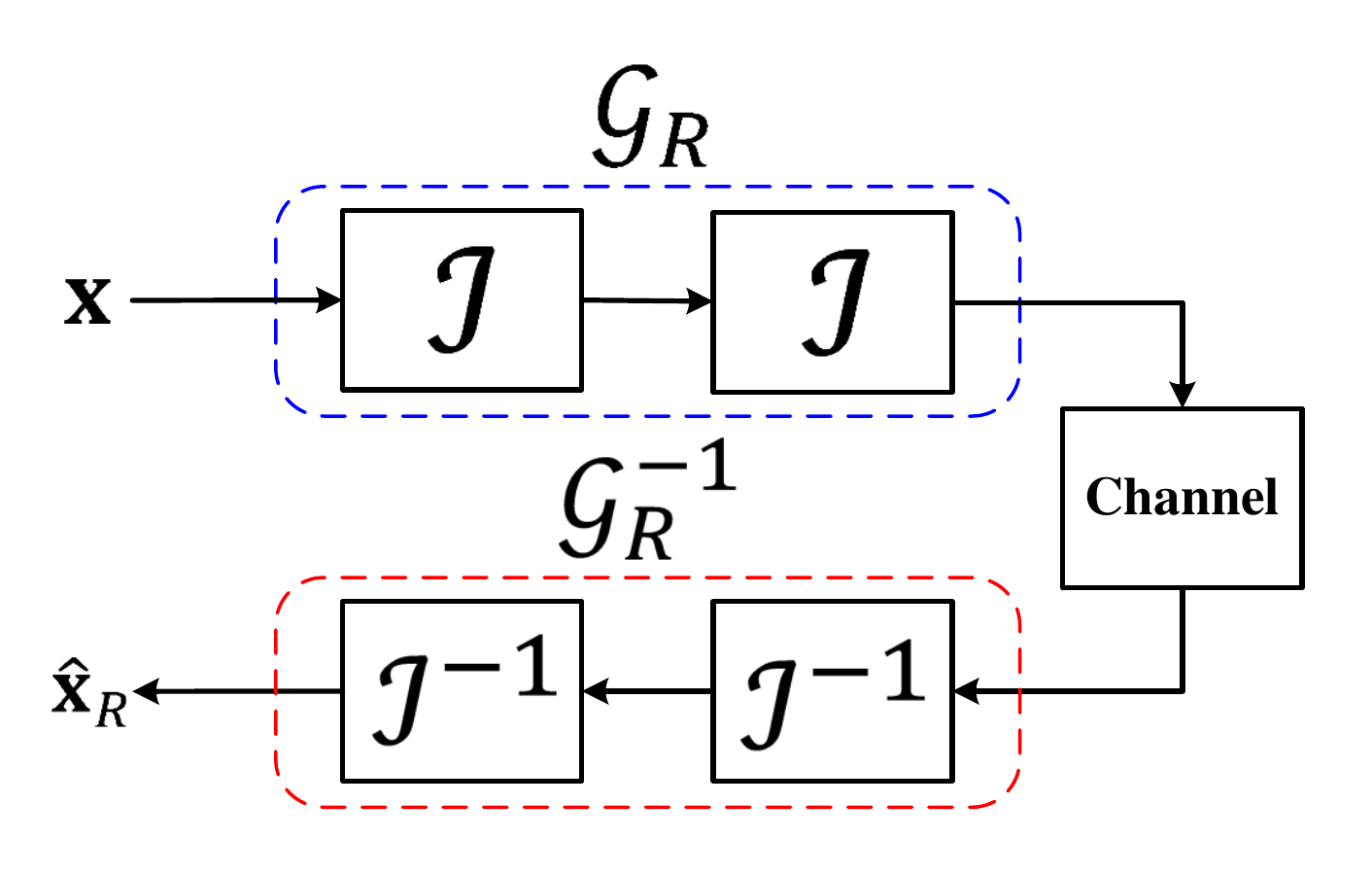}}
\hspace{0.5cm}%
%%----start of third subfigure----
\subfloat[]{
\label{fig:subfig:c} %% label for third subfigure
\includegraphics[width=5cm]{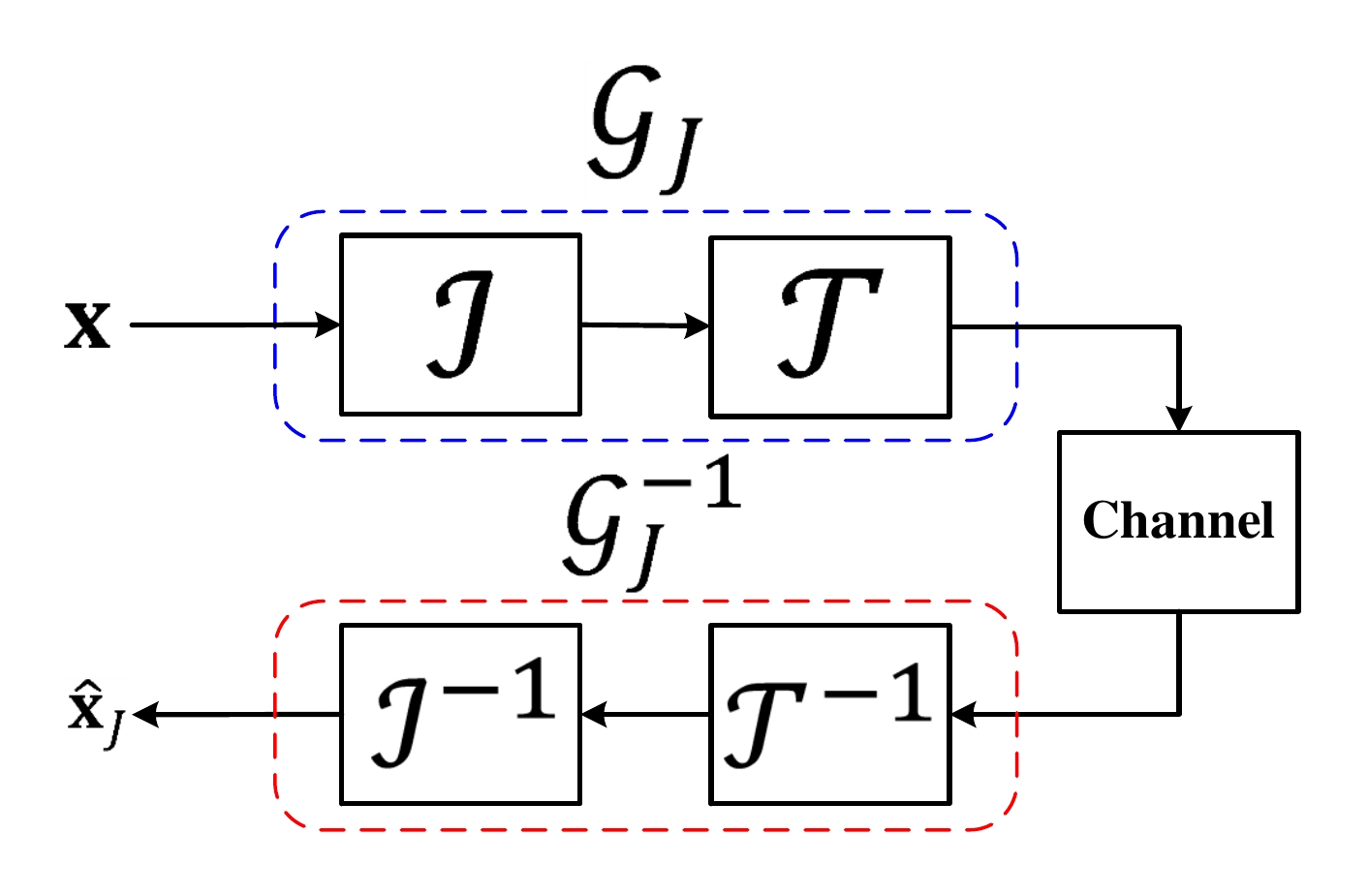}}
\caption{Simplified encoding and decoding models in compressive and conventional imaging systems. (a) The model of a compressive imaging system, (b) the model of a conventional raw camera, and (c) the model of a conventional camera equipped with a built-in compression stage.}
\label{ConventionalCompressiveModels} %% label for entire figure
\end{figure*}

Different mechanisms may be considered for designing application-specific compressive imagers \cite{duarte2008single, taimori2018adaptive}. Single-pixel camera is a physical implementation of an imaging system based on compressive sensing technology. In this camera, the natural image scene $\mathbf{X} \in \mathbb{R}^{h\times w}$ is first concentrated on a configurable array of mirrors, called Digital Micro-mirror Device (DMD), via a lens. The dimensions $h$ and $w$ represent the height and the width of the image $\mathbf{X}$, respectively. Based on compressed sensing theory, for each measurement cycle, only about 50\% of the analogy image projected onto DMD are selected by a predefined random pattern, modeled as the column vector $\mathbf{a}\in\mathbb{R}^{n}$, in which $n=h\times w$. Then, the randomly sampled image is collected on a photo-diode by another lens. The sensor signal corresponding to the modulated light is converted to a single discrete measurement $y_i, \forall i=1, 2, \cdots, m$. For the given number of measurements $m$, this cycle is repeated, which results in the measurement vector $\mathbf{y}\triangleq \mathbf{Ax}\in \mathbb{R}^m$, where $\mathbf{A}\in \mathbb{R}^{m\times n}$ represents a wide sensing matrix arranged by $\mathbf{A}=[\mathbf{{a}}_1^{\text T}, \mathbf{{a}}_2^{\text T}, \cdots, \mathbf{{a}}_m^{\text T}]^{\text{T}}$, i.e., we have usually $m\ll n$, so that $R_s = \frac{m}{n}$ is known as the sampling rate or compressive ratio. Finally, a decoder algorithm gets both the measurement vector $\mathbf{y}$ and the wide sensing matrix $\mathbf{A}$ and then exposes the image scene. The single-pixel camera recovers images using the ${\ell_1}$-norm convex minimization algorithm, also called Basis Pursuit (BP) in literatures as the following $\text{P}_{1, \delta}$ problem
%\[\mathbf{A}=\left( \begin{array}{c}
%\mathbf{{a}}_1^{\text T} \\
%\mathbf{{a}}_2^{\text T} \\
%\vdots  \\
%\mathbf{{a}}_m^{\text T} \end{array}
%\right),\]
\begin{equation}
\label{BasisPursuit}
\text{P}_{1, \delta}: \begin{array}{ccc}
{\mathop{\mathrm{min}}_{\boldsymbol{\mathrm{x}}} {\left\|\boldsymbol{\mathrm{x}}\right\|}_{{\ell }_1}\ } & \mathrm{s.t.} & {\left\|\boldsymbol{\mathrm{y}}-\boldsymbol{\mathrm{Ax}}\right\|}^2_{{\ell }_2}\le \delta,  \end{array}
\end{equation}
where  $\delta \ge \epsilon $ and  $\epsilon \triangleq {\left\|\boldsymbol{\mathrm{n}}\right\|}^2_{{\ell }_2}$ denotes the energy of the additive noise term $\mathbf{n}$. The symbol ${\parallel \cdot \parallel}_{\ell_p}$ also represents the ${\ell_p}$-norm.

The above process of imaging and representation of a digital image in CS technique can be modeled by two general block of the encoder $\mathcal{G}$ and the decoder $\mathcal{G}^{-1}$. Figure~\ref{CompressiveImagingModel} shows the process of coding and decoding of a compressive imaging system using some mathematical operators. Generally, the encoding process is done into a compressive camera. In the destination which contains computers or servers with hardware components more powerful than the encoder side, the decoding process is performed. The need to powerful hardware is due to the computational complexity of decoder algorithms which usually are based on iterative or convex optimization methods \cite{taimori2018adaptive}. Because compressive sensing performs a many-to-one transform in the encoding phase, the decoder is not exactly inverse system of the encoder, i.e. the function $\mathcal{G}$ is not invertible and $\mathcal{G}^{-1}\{\mathcal{G}(\mathbf{X})\}\neq \mathbf{X}$. Some information loss always exist, therefore, $\mathcal{G}^{-1}\{\mathcal{G}(\mathbf{X})\}= \widehat{\mathbf {X}}$, which can be exploited as an intrinsic artifact in compressive imaging forensics.

For the transmitter shown in Fig.~\ref{CompressiveImagingModel}, the operators $\mathcal{R}$, $\mathcal{F}$, $\mathcal{M}$, $\mathcal{Q}$, and $\mathcal{C}$ represent parallel raster scanner, sparsifying transform, sampling or measurement operator, quantizer, and coder, respectively. In the receiver, the blocks $\mathcal{R}^{-1}$, $\mathcal{F}^{-1}$, $\mathcal{M}^{-1}$, $\mathcal{Q}^{-1}$, and $\mathcal{C}^{-1}$ denote inverse parallel raster scanner, inverse sparsifying or densifying transform, reconstructor, dequantizer, and decoder, respectively. The task of the block $\mathcal{R}$ is to scan parallel-wise in the vertical direction of signal samples. In fact, this block converts a matrix to a vector, i.e. $\mathbf{x}=\mathcal{R}\{\mathbf{X}\}$, and its inverse system, i.e. $\mathcal{R}^{-1}$ is exactly available. This also forensically means the scan operation is loss-less and has not any trace. The block $\mathcal{F}$ contains a transform in which the signal in that domain is sparse. For instance, the transformations Fourier, Discrete Cosine Transform (DCT), wavelet, and Haar have such property. The inverse of these transformations are exactly at hand without information loss. In the operator $\mathcal{M}$, which can be modeled by the matrix $\mathbf{M}$, the signal is measured. For example, the measurement approach can be a random Gaussian matrix or random sampling \cite{candes2006robust}. In the block $\mathcal{M}$, information loss exist due to many-to-one conversion and hence the operator $\mathcal{M}^{-1}$ in not completely available. Factually, each measurement consists of a weighted linear combination of all signal samples or a combination of some random samples of the signal. The reconstruction of the signal is almost performed by iterative or convex optimization procedures such as \eqref{BasisPursuit}. The quantizer $\mathcal{Q}$ is a lossy system and leaves quantization errors in practice. The coder $\mathcal{C}$ is generally a loss-less operation such as Huffman or arithmetic codes and the inverse system is completely available. Here, we assume the communication channel or storage space is ideal, i.e. error-, noise-, and distortion-less, even though, in practice, channel coding/decoding are utilized, too.

To formulate the problem, the recovered image can be determined as
\begin{equation}
\label{RecoveredImage}
\widehat{\mathbf {X}}= \mathcal{G}^{-1}\mathcal{G}(\mathbf{X}),
\end{equation}
where the complete model is
\begin{equation}
\begin{array}{l}
\label{RecoveredImage_CompleteModel}
\widehat{\boldsymbol{\mathrm{X}}}=\\
\mathcal{R}^{-1}\left\{{\mathcal{F}}^{-1}\left[{\mathcal{M}}^{-1}\left({\mathcal{Q}}^{-1}\left({\mathcal{C}}^{-1}\left(\mathcal{C}\left(\mathcal{Q}\left(\mathcal{M}\left(\mathcal{F}\left(\mathcal{R}\left(\boldsymbol{\mathrm{X}}\right)\right)\right)\right)\right)\right)\right)\right)\right]\right\}.
\end{array}
\end{equation}

%\begin{equation}
%\label{RecoveredImage_CompleteModel}
%%\widehat{\mathbf {X}}= %\mathcal{R}^{-1}\{\mathcal{F}^{-1}[\mathcal{M}^{-1}(\mathcal{Q}^{-1}(\mathcal{C}^{-1}(\mathcal{C}(\mathcal{Q}(\mathcal{M}(\mathcal{F}(\mathcal{R}(\mathbf{X})))))))]\}.
%%\widehat{\boldsymbol{\mathrm{X}}}={\mathcal{R}}^{-1}\left\{{\mathcal{F}}^{-1}\left[{\mathcal{M}}^{-1}\left(\mathcal{M}\left(\mathcal{F}\left(\mathcal{R}\left(\boldsymbol{\mathrm{X}}\right)\right)\right)\right)\right]\right\}
%\widehat{\boldsymbol{\mathrm{X}}}=\\
%{\mathcal{R}}^{-1}\left\{{\mathcal{F}}^{-1}\left[{\mathcal{M}}^{-1}\left({\mathcal{Q}}^{-1}\left({\mathcal{C}}^{-1}\left(\mathcal{C}\left(\mathcal{Q}\left(\mathcal{M}\left(\mathcal{F}\left(\mathcal{R}\left(\boldsymbol{\mathrm{X}}\right)\right)\right)\right)\right)\right)\right)\right)\right]\right\}
%\end{equation}

\subsection{Forensically Simplified Imaging Models}

\subsubsection{Simplified Compressive Imaging Model}
As seen in~\eqref{RecoveredImage_CompleteModel}, the model of a compressive imaging system is complex. This fact harden forensic investigation of the system. For simplicity, we concentrate our focus on the measurement block $\mathcal{M}$ and ignore other sources of errors such as quantization noise. In the decoder phase, the operator $\mathcal{M}^{-1}$ for reconstructing the transmitted signal uses convex optimization methods such as the ${\ell_1}$-norm or nonlinear iterative algorithms. Applying such an operation leaves some artifacts such as high frequency oscillations and blurring. We employ these artifacts as forensic footprints for compressive imaging history identification. In this case, the simplified model of imaging is considered as follows
\begin{equation}
\label{RecoveredImage_SimplifiedModel}
%\widehat{\mathbf {X}}= %\mathcal{R}^{-1}\{\mathcal{F}^{-1}[\mathcal{M}^{-1}(\mathcal{Q}^{-1}(\mathcal{C}^{-1}(\mathcal{C}(\mathcal{Q}(\mathcal{M}(\mathcal{F}(\mathcal{R}(\mathbf{X})))))))]\}.
\widehat{\boldsymbol{\mathrm{X}}}={\mathcal{R}}^{-1}\left\{{\mathcal{F}}^{-1}\left[{\mathcal{M}}^{-1}\left(\mathcal{M}\left(\mathcal{F}\left(\mathcal{R}\left(\boldsymbol{\mathrm{X}}\right)\right)\right)\right)\right]\right\}.
\end{equation}
Since the blocks $\mathcal{R}$ and $\mathcal{R}^{-1}$ are loss-less, we can reduce~\eqref{RecoveredImage_SimplifiedModel} for the input vector $\mathbf{x}$ as
\begin{equation}
\label{RecoveredImage_MoreSimplerModel}
\widehat{\boldsymbol{\mathrm{x}}}={\mathcal{F}}^{-1}\left\{{\mathcal{M}}^{-1}\left[\mathcal{M}\left(\mathcal{F}\left(\boldsymbol{\mathrm{x}}\right)\right)\right]\right\}.
\end{equation}

In order to find forensically important traces, we model conventional and compressive imaging systems as a deconvolution problem. In this case, we seek a representation blurring kernel for each imaging system. The configuration of kernel function can be utilized for detecting the history of compressive imaging.

\begin{figure}[!t]
\centering
\includegraphics[width=5cm]{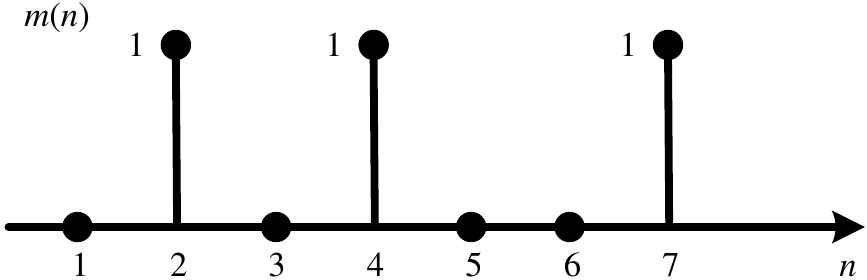}
\caption{A typical example of a random sampling mask with the length of 7.}
\label{CS_Example}
\end{figure}

Figure~\ref{ConventionalCompressiveModels} (a) depicts the reduced compressive imaging model. The input signal $\mathbf{x}$ maps into a domain via the transform $\mathcal{F}$ in which the signal is almost sparse. Then, by using the operator $\mathcal{M}$, the transformed signal is sampled. The operation $\mathcal{M}$ may be considered as a diagonal matrix where only a given percent of its main diagonal elements that their locations are selected randomly, are one and in other locations are zero. The percent of the main diagonal elements with the value one is known as the sampling rate.
\newtheorem{Example}{Example}
\begin{Example}[Random sampling as a special case of CS]
\label{Ex1}
Consider the random sampling mask $m(n)$ in Fig.~\ref{CS_Example} with the length $n=7$. This signal can be represented by the vector $\mathbf{m}={\left[m_1, m_2, \cdots, m_7\right]}^{\mathrm{T}}={\left[ \begin{array}{ccccccc}
0 & 1 & 0 & 1 & 0 & 0 & 1 \end{array}
\right]}^{\mathrm{T}}$. In this case, each row of the measurement matrix $\mathbf{M}$ consists of only a single 1 and all of the other entries are 0, i.e.
\begin{equation} \label{MeasurementMatrixExample}
{\mathbf{M}}=\left( \begin{array}{ccccccc}
{\rm 0} & {m_2} & {\rm 0} & {\rm 0} & {\rm 0} & {\rm 0} & {\rm 0} \\
{\rm 0} & {\rm 0} & {\rm 0} & {m_4} & {\rm 0} & {\rm 0} & {\rm 0} \\
{\rm 0} & {\rm 0} & {\rm 0} & {\rm 0} & {\rm 0} & {\rm 0} & {m_7} \\ \end{array}
\right).
\end{equation}
\end{Example}

\newtheorem{Lemma}{Lemma}
\begin{Lemma}[Blurring kernel of compressive imaging system]\label{Lem1}
Consider the compressive imaging model of Fig.~\ref{ConventionalCompressiveModels} (a), in which the recovered signal $\widehat{\boldsymbol{\mathrm{x}}}_C$ is determined as
\begin{equation}
\label{RecoveredCS_Signal}
\widehat{\boldsymbol{\mathrm{x}}}_C={\mathcal{G}}^{-1}_C{\mathcal{G}}_C\left(\boldsymbol{\mathrm{x}}\right).
\end{equation}
The encoding phase of compressive sensing is modeled as a linear under-determined system of equations. However, similar to almost all real world systems, the decoder is a nonlinear system, because the signal is generally recovered by nonlinear iterative algorithms. To be able to analyze the system behavior, suppose the decoding function be approximated by a Linear Shift-Invariant (LSI) system around the operating point. E.g., Taylor series expansion may be used for linearization \cite{ogata2004system}. In such a situation, the reconstructed signal $\widehat{x}_C(n)$ can be modeled as the convolution of the signal $x(n)$ with the unknown blur kernel $h_C(n)$, i.e. $\widehat{x}_C(n)=h_C(n)\ast x(n)$. Based on Toeplitz and circulant matrices~\cite{gray2006toeplitz}, this convolution operation can be rewritten in an algebraic vector-matrix form as
\begin{equation}
\label{RecoveredCS_Signal_vector_Matrix_Form}
\widehat{\boldsymbol{\mathrm{x}}}_C\approx{\boldsymbol{\mathrm{H}}}_C\boldsymbol{\mathrm{x}},
\end{equation}
where the matrix $\boldsymbol{\mathrm{H}}_C$ represents the blurring kernel of compressive imaging system. If the matrices $\mathbf{F}$, $\mathbf{M}$, and ${\widehat{\mathbf{M}}}^{-1}$ respectively denote the sparsifying transform, sampler, and the linearized approximate model of the system ${\mathcal{M}}^{-1}$, then
\begin{equation}
\label{CS_Kernel}
\boldsymbol{\mathrm{H}}_C\approx{\left(\boldsymbol{\widehat{\mathrm{M}}\mathrm{F}}\right)}^{-1}\left(\boldsymbol{\mathrm{MF}}\right).
\end{equation}
By defining the matrix $\widehat{\mathbf{A}}\triangleq{\widehat{\mathbf{M}}}\mathbf{F}$ as well as the sensing matrix $\boldsymbol{\mathrm{A}}\boldsymbol{\mathrm{\triangleq }}\boldsymbol{\mathrm{MF}}$, we have
\begin{equation}
\label{RecoveredCS_Signal_vector_Matrix_Form}
\boldsymbol{\mathrm{H}}_C\approx{{\widehat{\mathbf{A}}}}^{-1}\mathbf{A}.
%\widehat{\boldsymbol{\mathrm{x}}}_C\approx{\boldsymbol{\mathrm{H}}}_C\boldsymbol{\mathrm{x}}
\end{equation}
\end{Lemma}

\begin{Example}[CS recovery by minimizing the ${\ell_2}$-norm]
\label{Ex2}
A simple closed-form but inaccurate sparse recovery solution is the ${\ell_2}$-norm minimization, i.e. ${\boldsymbol{\mathrm{H}}}_C\approx{\boldsymbol{\mathrm{A}}}^{\dagger }\boldsymbol{\mathrm{A}}$, in which the matrix ${\boldsymbol{\mathrm{A}}}^{\dagger }\boldsymbol{\mathrm{\triangleq }}{\boldsymbol{\mathrm{A}}}^{\mathrm{T}}{\left(\boldsymbol{\mathrm{A}}{\boldsymbol{\mathrm{A}}}^{\mathrm{T}}\right)}^{-1}$ indicates the pseudo-inverse matrix for the under-determined system of equations.
\end{Example}

\subsubsection{Conventional Raw Imaging Model}
In cameras which deliver raw images, pixels are directly measured from CCD or CMOS sensors without any post-processing or compression on the sensed signal. This type of imaging is usually utilized for professional digital photography. To formulate the problem in this case, we extend the reduced CS model shown in Fig.~\ref{ConventionalCompressiveModels} (a) to conventional imaging systems. In the CS representation, the matrix $\mathbf{M}$ may be considered as a diagonal matrix, where only a given percent of its main diagonal elements that their locations have sparsely distributed, are 1 and in other entries are 0. The ratio of the number of main diagonal elements with the value 1 to the length of signal gives the sampling rate $R_s = {\frac{{\parallel \text{diag}(\mathbf{M}) \parallel}_{\ell_0}}{n}}\times 100$ in percent.

If $R_s \rightarrow 100\%$, then the CS model tends to a conventional sensing. In other words, a conventional sensing model can be interpreted as the complete measurement of the vector containing signal $\mathbf{x}$, without any transformation from the sample domain to a sparsifying domain. To demonstrate this, substituting $\mathbf{M}=\mathbf{I}$ in \eqref{CS_Kernel} gives

$\boldsymbol{\mathrm{H}}_C={\left(\boldsymbol{\mathrm{IF}}\right)}^{-1}\left(\boldsymbol{\mathrm{IF}}\right) \Rightarrow \boldsymbol{\mathrm{H}}_C={\boldsymbol{\mathrm{F}}}^{\mathrm{-1}}\underbrace{{\boldsymbol{\mathrm{I}}}^{\mathrm{-1}}\boldsymbol{\mathrm{I}}}_{\boldsymbol{\mathrm{=}}\boldsymbol{\mathrm{I}}}\boldsymbol{\mathrm{F}}$,

\noindent where yields $\boldsymbol{\mathrm{H}}_C=\mathbf{F}^{-1}\mathbf{F}$, and ultimately $\boldsymbol{\mathrm{H}}_C=\mathbf{I}$. Figure~\ref{ConventionalCompressiveModels} (b) shows the raw imaging system. In this scheme, the matrices transform and measurement are the same and equal to the identity matrix, i.e. $\mathcal{I}\mathrm{=}\boldsymbol{\mathrm{I}}$. In the receiver, just the main signal is usable.
\begin{Example}[Sampling matrix for conventional imaging]
\label{Ex3}
As mentioned above, the the measurement matrix \eqref{MeasurementMatrixExample} can be rewritten in an equivalent form as
%\begin{equation} \label{MeasurementMatrixInSquareFormExample}
%{\mathbf{M}}=\left( \begin{array}{ccccccc}
%{m_1} & {\rm 0} & {\rm 0} & {\rm 0} & {\rm 0} & {\rm 0} & {\rm 0} \\
%{\rm 0} & {m_2} & {\rm 0} & {\rm 0} & {\rm 0} & {\rm 0} & {\rm 0} \\
%{\rm 0} & {\rm 0} & {m_3} & {\rm 0} & {\rm 0} & {\rm 0} & {\rm 0} \\
%{\rm 0} & {\rm 0} & {\rm 0} & {m_4} & {\rm 0} & {\rm 0} & {\rm 0} \\
%{\rm 0} & {\rm 0} & {\rm 0} & {\rm 0} & {m_5} & {\rm 0} & {\rm 0} \\
%{\rm 0} & {\rm 0} & {\rm 0} & {\rm 0} & {\rm 0} & {m_6} & {\rm 0} \\
%{\rm 0} & {\rm 0} & {\rm 0} & {\rm 0} & {\rm 0} & {\rm 0} & {m_7} \\ \end{array}
%\right),
%\end{equation}
\begin{equation} \label{MeasurementMatrixInSquareFormExample}
\mathbf{M}=\text{diag}(m_1, m_2, \cdots, m_7),
\end{equation}
for which $R_s =42.86\%$. However, for conventional sensing, the matrix shown in \eqref{MeasurementMatrixInSquareFormExample} converts to a identity matrix, i.e. $\mathbf{M}={\mathbf{I}}_{7\times 7}$ with $R_s =100\%$.
\end{Example}

\begin{Lemma}[Blurring kernel of conventional raw imaging]\label{Lem2}
Consider the traditional raw imaging model of Fig.~\ref{ConventionalCompressiveModels} (b), in which the recovered signal $\widehat{\boldsymbol{\mathrm{x}}}_R$ is determined as
\begin{equation}
\label{RecoveredRaw_Signal}
\widehat{\boldsymbol{\mathrm{x}}}_R={\mathcal{G}}^{-1}_R{\mathcal{G}}_R\left(\boldsymbol{\mathrm{x}}\right).
\end{equation}
Based on the vector-matrix representation, the raw imaging model can be described as
\begin{equation}
\label{RecoveredRaw_Signal_vector_Matrix_Form}
\widehat{\boldsymbol{\mathrm{x}}}_R={\boldsymbol{\mathrm{H}}}_R\boldsymbol{\mathrm{x}},
\end{equation}
where the matrix $\boldsymbol{\mathrm{H}}_R={\left(\boldsymbol{\mathrm{II}}\right)}^{-1}\left(\boldsymbol{\mathrm{II}}\right)$ represents the blurring kernel of the raw imaging system. This results in
\begin{equation}
\label{RecoveredRaw_Signal_vector_Matrix_Form}
%\boldsymbol{\mathrm{H}}_R={\left(\boldsymbol{\mathrm{II}}\right)}^{-1}\left(\boldsymbol{\mathrm{II}}\right),
%\widehat{\boldsymbol{\mathrm{x}}}_C\approx{\boldsymbol{\mathrm{H}}}_C\boldsymbol{\mathrm{x}}
\boldsymbol{\mathrm{H}}_R=\boldsymbol{\mathrm{I}},
\end{equation}
which $\mathbf{I}$ denotes the identity matrix. Based on this model, the original signal is perfectly reconstructible in the decoder. Therefore, we have ${\widehat{\boldsymbol{\mathrm{x}}}}_R=\boldsymbol{\mathrm{x}}$.
\end{Lemma}

\subsubsection{Conventional Imaging Plus Compression}
The most of ordinary cameras such as those embedded in today smart cell phones, at first, grab an image and then compress the content. Similar to raw imaging, the total number of pixels are first measured. This means that compression is not performed at sensor level. But, after some optional post-processing operations, the signal is coded using a built-in or external source coder such as JPEG2000 in order to reduce the bit rate. In Fig.~\ref{ConventionalCompressiveModels} (c), we plan this kind of imaging, where total compression operation has modeled by the operator $\mathcal{T}$. For JPEG2000 compression standard, the main encoding steps include the transform to wavelet domain, quantization, and binary coding, which are performed sequentially. The parameter of the compression ratio, $R_c\triangleq\frac{\text{The input image size}}{\text{The output compressed size}}$, in the quantizer controls the required bit rate, which is a real number more than or equal to 1. In the decoder $\mathcal{T}^{-1}$, the reverse corresponding operations are applied, i.e. binary decoding, dequantization, and inverse wavelet transform, respectively. Based on this model, the main intrinsic fingerprint of JPEG2000 is the nonlinear quantizer system in the encoder and other operations are linear and losslessly decodable.
\begin{Lemma}[Conventional imaging plus compression kernel]\label{Lem3}
Consider the traditional raw imaging plus compression model of Fig.~\ref{ConventionalCompressiveModels} (c), in which the recovered signal $\widehat{\boldsymbol{\mathrm{x}}}_J$ is determined as
\begin{equation}
\label{RecoveredRawPlusCompression_Signal}
\widehat{\boldsymbol{\mathrm{x}}}_J={\mathcal{G}}^{-1}_J{\mathcal{G}}_J\left(\boldsymbol{\mathrm{x}}\right).
\end{equation}
Suppose the model of JPEG2000 compressor be approximated by a LSI system, then, the reconstructed signal in the form of vector-matrix representation can be formulated as
\begin{equation}
\label{RecoveredRawPlusCompression_Signal_vector_Matrix_Form}
\widehat{\boldsymbol{\mathrm{x}}}_J\approx{\boldsymbol{\mathrm{H}}}_J\boldsymbol{\mathrm{x}}.
\end{equation}
The matrix $\boldsymbol{\mathrm{H}}_J\approx{\left(\mathbf{TI}\right)}^{-1}\left({\widehat{\mathbf{T}}\mathbf{I}}\right)$ represents the blurring kernel of raw imaging together with compression, in which the matrices $\widehat{\mathbf{T}}$ and ${\mathbf{T}}^{-1}$ denote the linearized approximate model of the compressor $\mathcal{T}$ and the linear decompressor $\mathcal{T}^{-1}$, respectively. The blurring kernel is simplified as
\begin{equation}
\label{RecoveredRawPlusCompression_Signal_vector_Matrix_Form}
\boldsymbol{\mathrm{H}}_J\approx{\mathbf{T}}^{-1}\widehat{\mathbf{T}}.
%\widehat{\boldsymbol{\mathrm{x}}}_C\approx{\boldsymbol{\mathrm{H}}}_C\boldsymbol{\mathrm{x}}
\end{equation}
In this model, if $R_c \rightarrow 1$, then $\widehat{\mathbf{T}}=\mathbf{T}=\mathbf{I}$, which yields $\boldsymbol{\mathrm{H}}_J=\mathbf{I}$.
\end{Lemma}

\section{Identification Methodology}
\label{IdentificationMethodology}
\begin{figure}[!t]
\centering
\includegraphics[width=\linewidth]{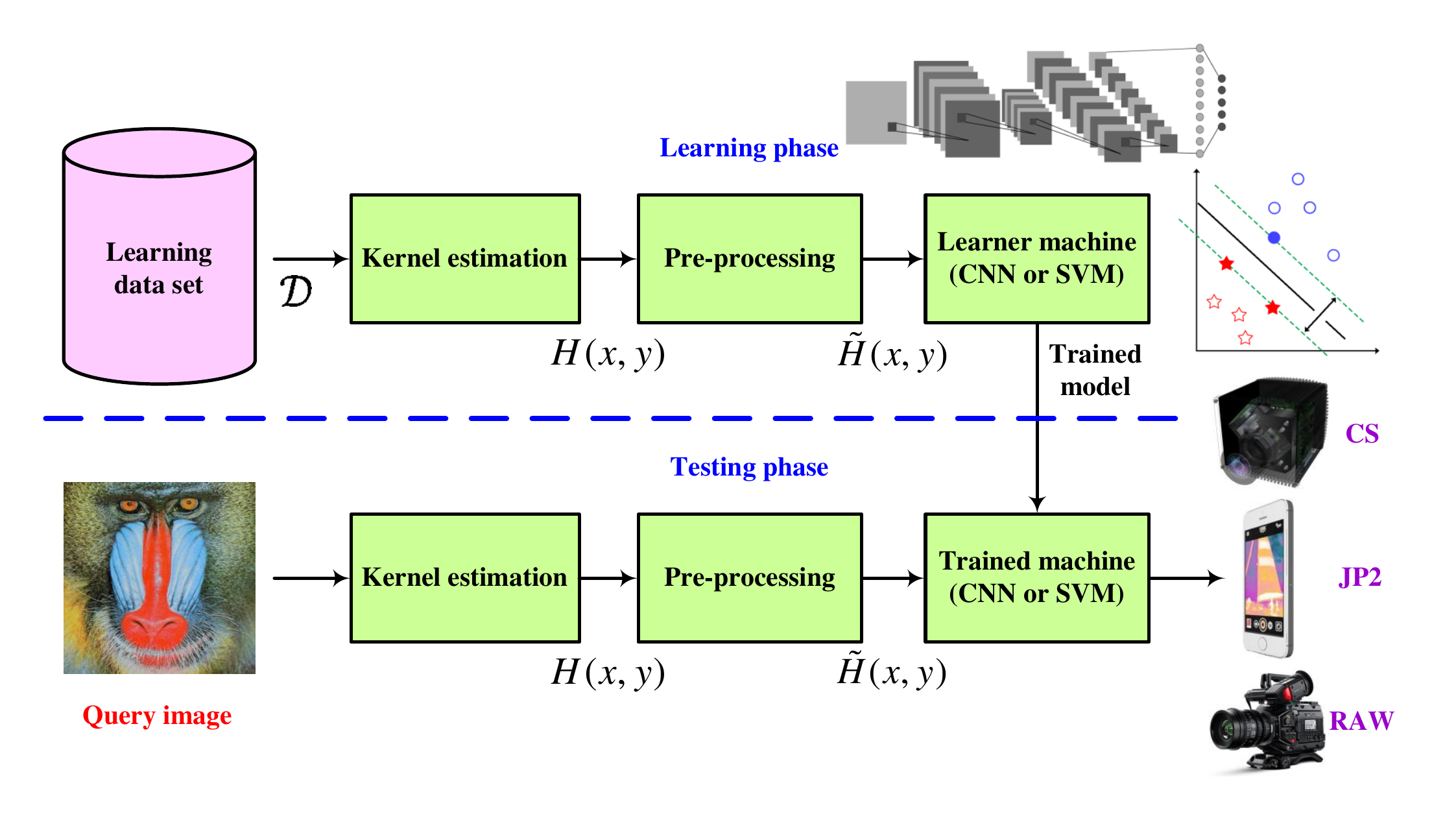}
\caption{The proposed flowchart for forensic identification of compressive imaging.}
\label{ProposedFlowchart}
\end{figure}

In Lemmas 1, 2, and 3, we determined an estimate of blurring kernels for different imaging systems. The characteristics of these kernels can be exploited as a trace for identification of the history of compressive imaging. In the following theorem, we first demonstrate that the kernel functions of various imaging systems are discriminative. Then, we present a learning-based approach to train discriminant kernels for detecting compressive imaging. Figure~\ref{ProposedFlowchart} depicts the suggested scheme for compressive imaging identification.
\newtheorem{Theorem}{Theorem}
\begin{Theorem}[Discriminability of kernels in imaging systems]\label{Theor}
Let different imaging systems be modeled as $\widehat{\boldsymbol{\mathrm{x}}}_i\approx{\boldsymbol{\mathrm{H}}}_i\boldsymbol{\mathrm{x}}, \forall i\in \{C, J, R\}$, in which the vectors $\boldsymbol{\mathrm{x}}$, $\widehat{\boldsymbol{\mathrm{x}}}_i$, and the matrix $\mathbf{H}_i$ denote the latent original sharp signal, the output signal of the imaging system $i^{\text{th}}$, and blur kernel for the imager $i^{\text{th}}$, respectively. And, the set $\{C, J, R\}$ represents the compressive, conventional raw plus JPEG2000 compression, and conventional raw imaging systems, respectively. Then, blur kernels of compressive $\mathbf{H}_C$, conventional raw $\mathbf{H}_R$, and raw plus compression $\mathbf{H}_J$ imaging systems are discriminant.
\end{Theorem}

\begin{IEEEproof}
Lemmas 1, 2, and 3 demonstrate that the blurring kernels of compressive, conventional raw, and raw plus compression imaging systems are determined by different matrices $\boldsymbol{\mathrm{H}}_C\approx{{\widehat{\mathbf{A}}}}^{-1}\mathbf{A}$, ${\boldsymbol{\mathrm{H}}}_R=\boldsymbol{\mathrm{I}}$, and $\boldsymbol{\mathrm{H}}_J\approx{\mathbf{T}}^{-1}\widehat{\mathbf{T}}$, respectively. Accordingly, the blurring kernel of various imaging models are discriminant, so that  ${\boldsymbol{\mathrm{H}}}_i{\boldsymbol{\mathrm{\neq }}\boldsymbol{\mathrm{H}}}_j, \forall i,j\in \{C, J, R\}, i\neq j$.
\end{IEEEproof}

\subsection{Blurring Kernel Estimation}
\begin{figure}[!t]
\centering
%%----start of first subfigure----
\subfloat[]{
\label{fig:subfig:a} %% label for first subfigure
\includegraphics[width=2.4cm]{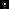}}
\hspace{0.5cm}%
%%----start of second subfigure----
\subfloat[]{
\label{fig:subfig:b} %% label for second subfigure
\includegraphics[width=2.4cm]{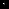}}
\hspace{0.5cm}%
%%----start of third subfigure----
\subfloat[]{
\label{fig:subfig:c} %% label for third subfigure
\includegraphics[width=2.4cm]{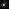}}
\caption{The estimated blur kernels for (a) a compressively sensed image, (b) a traditionally sensed image having compression, and (c) a conventional raw image.}
\label{KernelImages} %% label for entire figure
\end{figure}

As shown in Theorem 1, blurring kernels of different imaging systems are discriminative. In order to estimate the kernel $\mathbf{H}_i\in {\mathbb{R}}^{a\times b}, \forall i\in \{C, J, R\}$, we model the issue as a inverse problem. Given an image of an imaging system, the problem is to reconstruct original scene. Deconvolution is known as a paradigm of inverse problems~\cite{proakis2006digital}. Various direct and indirect deconvolution methods are available in order to estimate a degradation kernel and a deblurred version of an image such as Wiener filter and Richardson-Lucy algorithm \cite{levin2011understanding, schuler2016learning}. Here, we employed the well-known modified Maximum-A-Posterior (MAP)-based deconvolution algorithm proposed in~\cite{levin2011efficient}. By using EM optimization, this algorithm iterates alternatively between two steps, one for solving a latent sharp image and another for finding a blur kernel. In blind deconvolution, kernel dimensions are generally much less than image dimensions \cite{bahrami2015blurred}. The shape of a degradation kernel is also considered as a square with the length $a$. Figure~\ref{KernelImages} illustrates instances of the estimated blur kernel for three imaging systems. The difference among kernels can be seen with the naked eye.

\subsection{Learning Kernels}
Estimated 2-D kernels can be used to discriminate different imaging systems. To do this, we propose a supervised learning-based approach to train a learner machine from kernels characteristics. For training, different machine learning approaches can be employed. Here, we have utilized the  state-of-the-art Convolutional Neural Network (CNN) and Support Vector Machine (SVM) methods for learning.

\subsubsection{CNN Learner Machine}
CNN is one of well-known algorithms in deep machine learning, which is generally used for detection and identification of objects in images and videos \cite{goodfellow2016deep}. The way of thinking in deep CNN is different from approaches use feature engineering. Unlike feature engineering methods which features are manually extracted by domain-specific experts, CNN determines characteristics in an automatic manner directly from image pixels. The structure of a CNN consists of two building blocks. The first block contains feature detection layers and the second block includes classification layers. In the feature detection block, useful features are extracted by filtering, thresholding, and nonlinear down-sampling. These operations are applied by using a successive set of processing steps on input image pixels including convolution, Rectified Linear Unit (ReLU), and pooling, respectively. In the classification stage, the vectorized features obtained from the first block are fed to a fully-connected network such as Multi Layer Perceptrons (MLPs) to train weights and biases of the network from learning data. CNN employs softmax activation function to predict output probabilities of classes. Here, we considered two modes for training CNN, one idea for learning from images corresponding to blur kernels and another for learning from images pixels of various imaging systems.

In the pre-process of CNN learning/testing, we first normalize kernel values in the range $[0, 1]$ as
\begin{equation}
\label{KernelNormalization}
\tilde{H}(x, y)=\frac{H(x, y)-\min{(\mathbf{H})}}{\max{(\mathbf{H})}-\min{(\mathbf{H})}}.
\end{equation}
Then, normalized values are converted into 8-bit unsigned integers in the range $[0, 255]$. This pre-processed kernel is similar to an gray-level image and can be used for training (See three representative images in Fig.~\ref{KernelImages}). We examined different topologies for learning CNN and selected the one structure that maximizes training accuracy. Based on this constraint, the utilized sub-optimal architecture is the $a\times b$ input kernel image, $20$ convolutional filters of dimensions $5\times 5$, the pool size of $2\times 2$ without overlapping, and the fully-connected neural network with 3 outputs. The optimization procedure used for training is stochastic gradient descent algorithm with momentum.

For CNN learning from image pixels, the kernel estimation and the pre-processing step shown in Fig.~\ref{ProposedFlowchart} are not applied and images of learning data set are directly utilized to automatically extract features. In this straightforward training mode, we used a sub-optimal structure including the input gray-level image set of size $w\times h$, $30$ filters of dimensions $3\times 3$, the pool size of $2\times 2$ without overlapping, and the fully-connected network with 3 outputs.

\subsubsection{SVM Learning}
SVM is a supervised classifier originally designed for binary classification problems \cite{wang2005support}. In order to find optimal separating line, plane, or hyper-plane in the feature space, SVM solves a convex optimization problem to maximize a margin function which is determined by a subset of training samples, so-called support vectors. Main configurations of SVMs are linear SVMs with hard and soft margins, and nonlinear SVMs with different kernel functions. By using strategies such as one vs one and one vs all, SVM classifier is extendable for multi-class classification problems \cite{rocha2014multiclass}. Here, for our 3-class classification problem, we utilized the one vs one division strategy and SVMs with linear kernel.

The pre-process step for SVM learner includes normalization of features in the range $[0, 1]$. To do this, we first consider an individual Gaussian distribution for each feature, in such a way that $\mathcal{N}({\mu}_{\lowercase{i}}, \sigma_{\lowercase{i}}), \forall i \in [1, a\times b]$. Then, we obtain the parameters sample mean and standard deviation for each feature from training data. At the end, the normalization for a training/testing sample is done as
\begin{equation}
\label{FeatureNormalization}
{\widetilde{{\mathbf h}}}{\mathbf =}\left({{\mathbf h}}{\mathbf -}{\boldsymbol \mu }\right)\odiv {\boldsymbol \sigma },
\end{equation}
where the vectors ${\mathbf h}$ and ${\widetilde{{\mathbf h}}}$ represent vectorized versions of matrices $\mathbf{H}$ and $\tilde{\mathbf{H}}$, respectively. So, we have ${\boldsymbol \mu }=[{\mu}_{1}, {\mu}_{2}, \cdots, {\mu}_{\lowercase{a\times b}}]^{\text{T}}$ and ${\boldsymbol \sigma }=[{\sigma}_{1}, {\sigma}_{2}, \cdots, {\sigma}_{\lowercase{a\times b}}]^{\text{T}}$. The symbol $\odiv $ denotes the entry-by-entry division.

\section{Numerical Experiments}
\label{NumericalExperiments}
Simulations were done in MATLAB environment and run on an Intel Core i7 2.2GHz laptop with 8GB RAM. For implementing SVM, we used LIBSVM toolbox \cite{chang2011libsvm}.

\begin{figure*}[!t]
\centering
%%----start of first subfigure----
\subfloat[]{
\label{fig:subfig:a} %% label for first subfigure
\includegraphics[width=5.55cm]{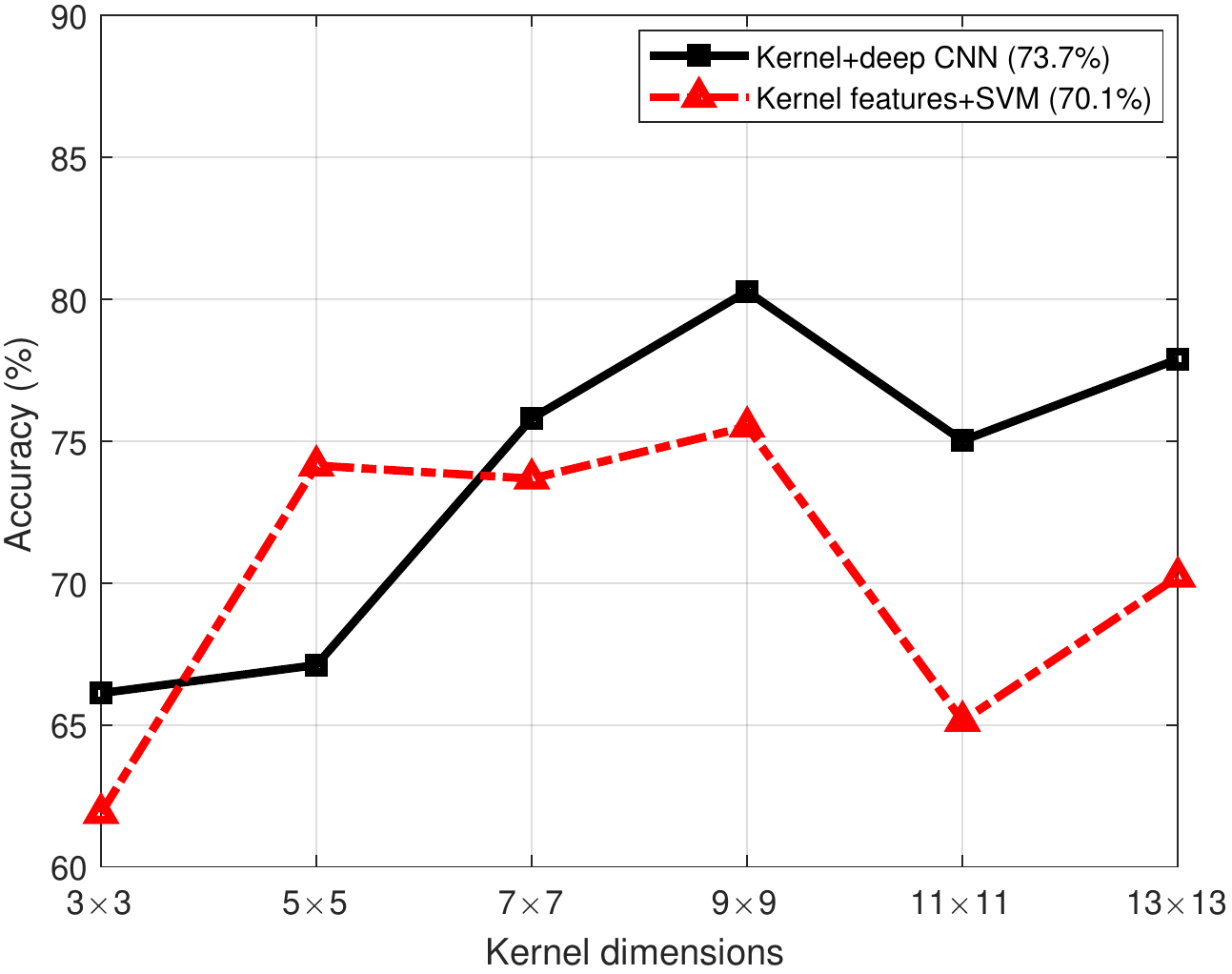}}
\hspace{0.5cm}%
%%----start of second subfigure----
\subfloat[]{
\label{fig:subfig:b} %% label for second subfigure
\includegraphics[width=5.55cm]{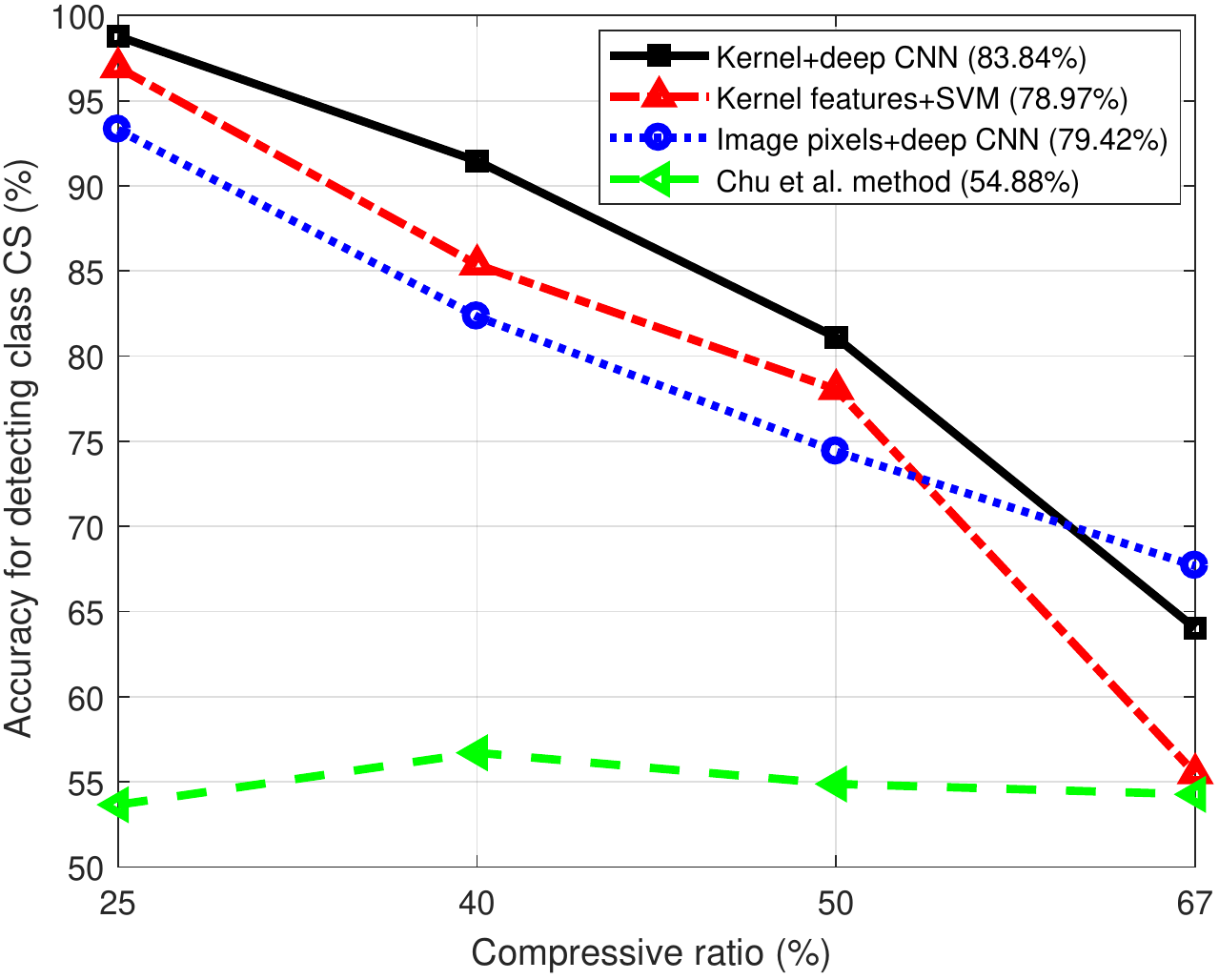}}
\hspace{0.5cm}%
%%----start of third subfigure----
\subfloat[]{
\label{fig:subfig:c} %% label for third subfigure
\includegraphics[width=5.55cm]{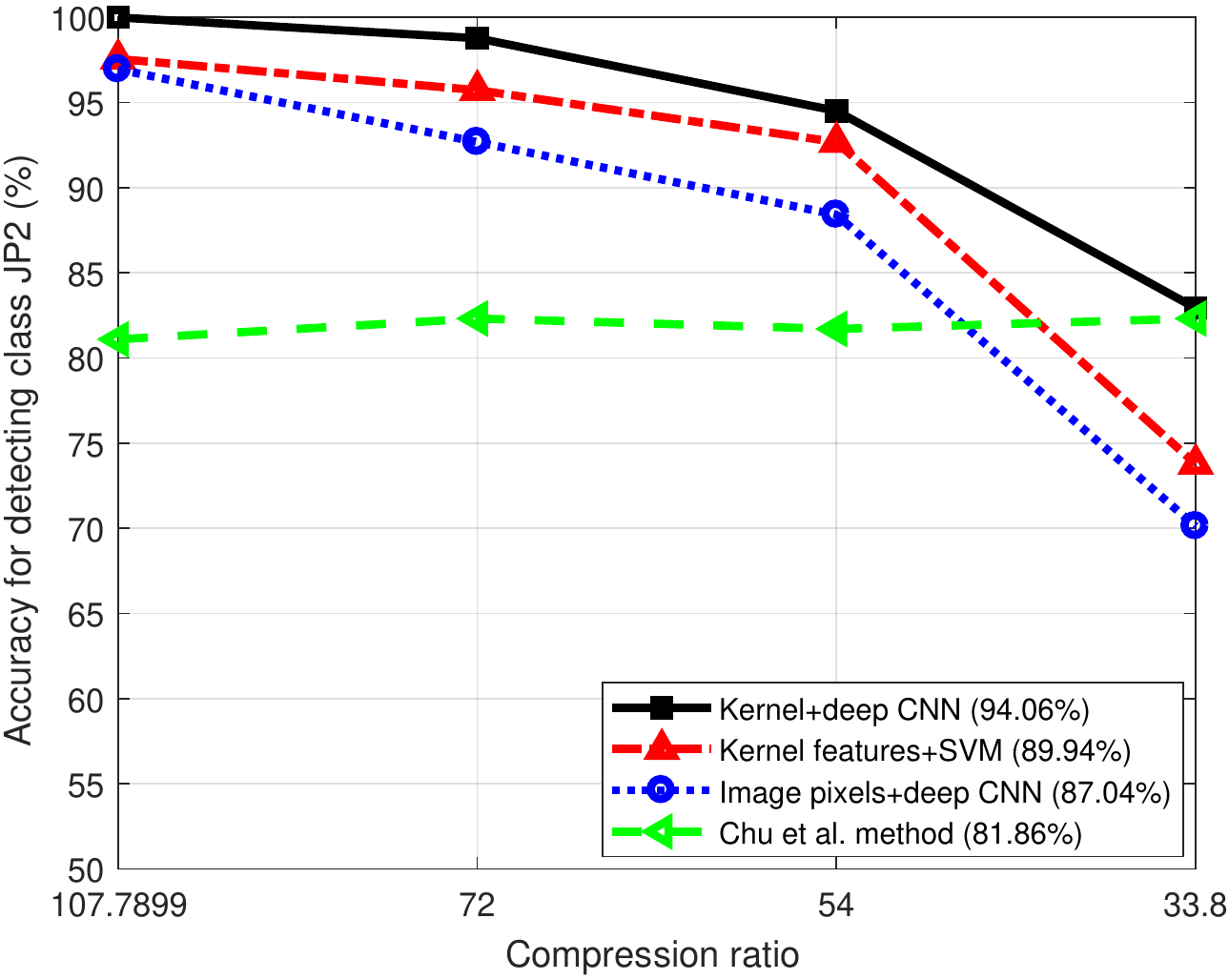}}
\caption{(a) The accuracy of proposed methods for different kernel dimensions, (b) the performance of detecting class CS under various compressive ratios, and (c) the performance of detecting class JP2 under various compression ratios.}
\label{PerEvaluations} %% label for entire figure
\end{figure*}

\subsection{Experimental Setups}
In experiments, we used the standard image set of Never-compressed Color Image Database (NCID) including $5150$ raw images with the dimensions $256\times 256$ and BMP file format~\cite{liu2011neighboring}. To create our experimental image set, we first divided NCID images into $w\times h=128\times 128$ non-overlapping blocks, and then randomly selected $N_r=655$ images patches among them. This yields the initial data set $\mathcal{D}_I=\{\mathbf{x}^i\}_{i=1}^{N_r}$.

We done experiments under different settings of imaging systems. The sampling rate in compressive sensing technology is usually set at a rate less than $50\%$ of the signal length~\cite{chu2015compressive}. To this intend, the sampling rates in the set $\mathcal{S}=\{25\%$, $40\%$, $50\%$, $67\%\}$ were considered for generating different compressively sensed images based on the encode/decoder designed in single-pixel camera~\cite{duarte2008single}. Therefore, in simulations, we used Gaussian measurements and BP algorithm for CS sampling and recovery, respectively. According to the above sampling rates, the total number of CS images was $N_r\cdot |\mathcal{S}|=2620$ images defined by the data set $\mathcal{D}_C\triangleq\{\widehat{\mathbf{x}}_C^i\}_{i=1}^{N_r\cdot|\mathcal{S}|}$. The symbol $|\cdot|$ denotes the cardinality of a set.

To create JP2 data set, we adjusted the parameter of compression ratio, $R_c$, in MATLAB JPEG2000 coder similar to the mechanism utilized in~\cite{chu2015compressive}. For a fair comparison, the expectation in this regime satisfies $\mathbb{E}\{\text{PSNR}|J\}\approx\mathbb{E}\{\text{PSNR}|C\}, \forall R_s \in \mathcal{S}$, where for each $R_s$, we have $\mathbb{E}\lbrace \text{PSNR}|k\rbrace \triangleq\frac{10}{N_{r}}\sum_{i=1}^{N_{r}}{\log
_{10}(\frac{M^{2}}{\frac{1}{n}\Vert \mathbf{x}^{i}-\mathbf{\hat{x}}_k^i\Vert
_{\ell_{2}}^{2}})}, \forall k\in \lbrace C, J\rbrace $.
\noindent The variable PSNR stands for Peak Signal to Noise Ratio and the parameter $M$ is the maximum value of a pixel, which is $255$ for an $8$-bit unsigned integer image. In our experiments, the compression ratio set corresponding to the set $\mathcal{S}$ was $\mathcal{P}=\{107.7899, 72, 54, 33.8\}$. This process results in $N_r\cdot |\mathcal{P}|=2620$ JP2 images of different compression ratio defined by the set $\mathcal{D}_J\triangleq\{\widehat{\mathbf{x}}_J^i\}_{i=1}^{N_r\cdot|\mathcal{P}|}$. The wavelet basis employed for creating both CS and JP2 images was biorthogonal 4.4 with 4 levels of decomposition.

In order to provide raw image set $\mathcal{D}_R$, in addition to the database $\mathcal{D}_I$, we randomly chosen other $1965$ images from initial $128\times 128$ patches without replacement defined by the set $\mathcal{D}_O$. Then, we collected them into the database $\mathcal{D}_R\triangleq\{\mathbf{x}^i\}_{i=1}^{|\mathcal{D}_I|+|\mathcal{D}_O|}$ including $|\mathcal{D}_I|+|\mathcal{D}_O|=2620$ images. In this case, the number of data for each class in the set $\{C, J, R\}$ is balanced. Generally speaking, learning from balanced classes leads to an appropriate performance \cite{he2009learning}.

The final database consists of the union of all databases, i.e. $\mathcal{D}\triangleq \bigcup_{i\in \{C, J, R\}}{\mathcal{D}_i}$, in which the total number of images is $|\mathcal{D}|=7860$. We exploited the hold-out cross validation mechanism, for which the first $75\%$ of data of each sub-class were selected as the learning set $\mathcal{D}_L$ and others for testing \cite{fukunaga2013introduction}. Hence, $\frac{\lfloor0.75N_r\rceil\cdot|\mathcal{D}|}{N_r}=5892$ images were chosen for training and the remainder $1968$ images for testing. The symbol $\lfloor\cdot\rceil$ denotes the nearest integer function. For evaluating abilities of different approaches in performance, we have reported metrics determined from confusion table.

\subsection{Evaluation under Different Kernel Sizes}
The length of estimated kernel is a parameter that may affect a restoration process. The goal of this experiment is to evaluate the accuracy of the proposed identification methods under a rational range of kernel length from $a=3$ to $13$. Figure~\ref{PerEvaluations} (a) depicts the curve of kernel dimensions vs accuracy for different learner systems. The envelope of curves shows the accuracy grows with increasing the kernel dimensions up to a desired point. After this size, the performance decreases, which is due to redundant information added for training. Hence, for the next experiments of this paper, we fixed the kernel length on a sub-optimal size that maximizes the accuracy constraint, i.e. $a^{*}=9$ for both deep CNN and SVM learner machines, as is clear in Fig.~\ref{PerEvaluations} (a). In the legend of Fig.~\ref{PerEvaluations} (a), the value in parentheses shows the average accuracy on all kernel dimensions for both CNN- and SVM-based methods, which infers the superiority of the CNN-based learner machine with the average accuracy of $73.7\%$.
In the case of kernel+deep CNN, we set the maximum number of epochs and the initial learning rate equal to $30$ and $0.0003$ in MATLAB, respectively. These parameter are respectively $30$ and $0.0001$ for the image pixels+deep CNN mode, too.

%\begin{figure}[!t]
%\centering
%\includegraphics[width=\linewidth]{PlotOfKernelSize_vs_Accuracy.pdf}
%\caption{The curve of kernel dimensions vs accuracy for the proposed methods.}
%\label{PlotOfKernelSize_vs_Accuracy}
%\end{figure}

\subsection{Discriminability Analysis by Features Visualization}
\begin{figure*}[!t]
\centering
%%----start of first subfigure----
\subfloat[]{
\label{fig:subfig:a} %% label for first subfigure
\includegraphics[width=5.55cm]{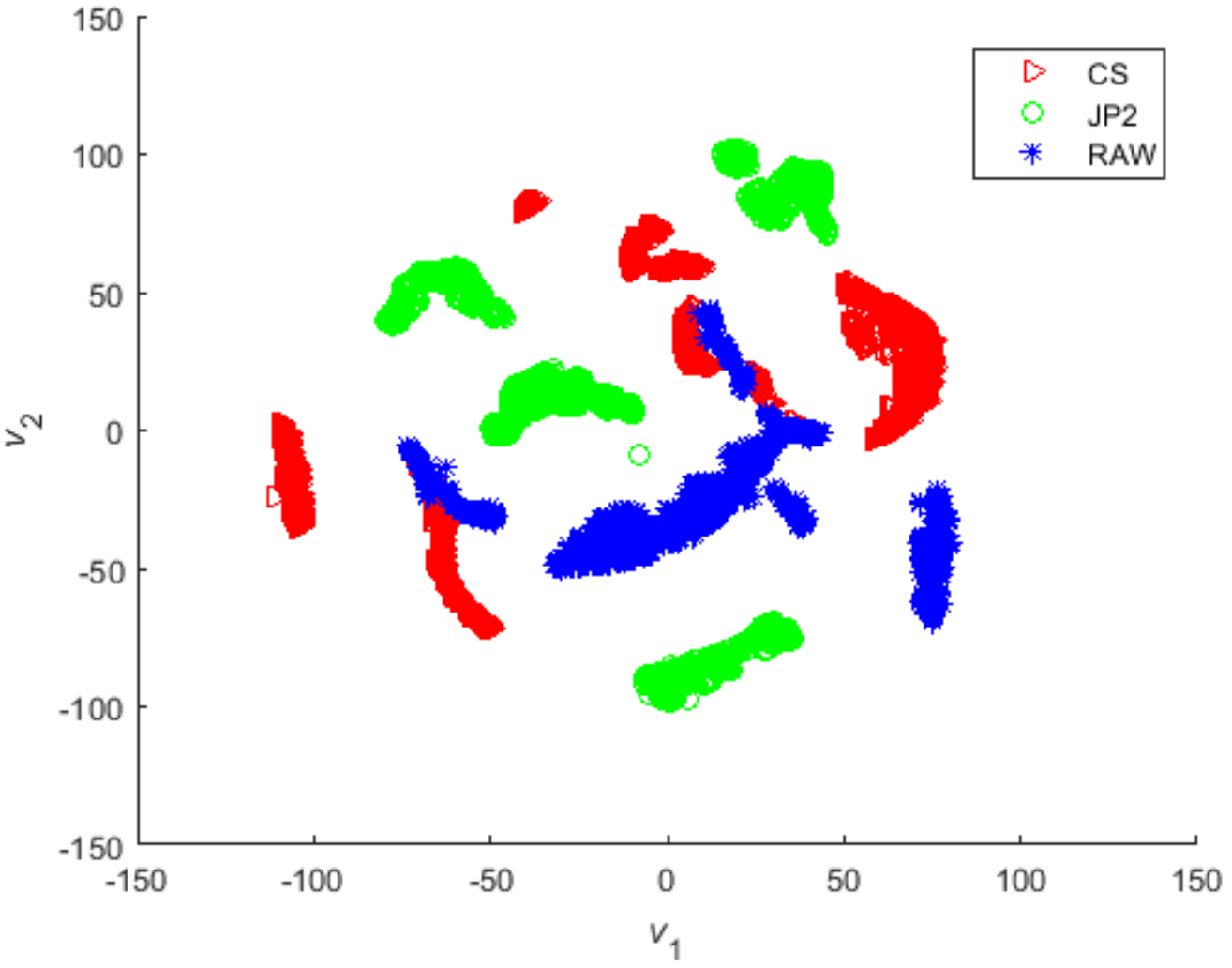}}
\hspace{0.5cm}%
%%----start of second subfigure----
\subfloat[]{
\label{fig:subfig:b} %% label for second subfigure
\includegraphics[width=5.55cm]{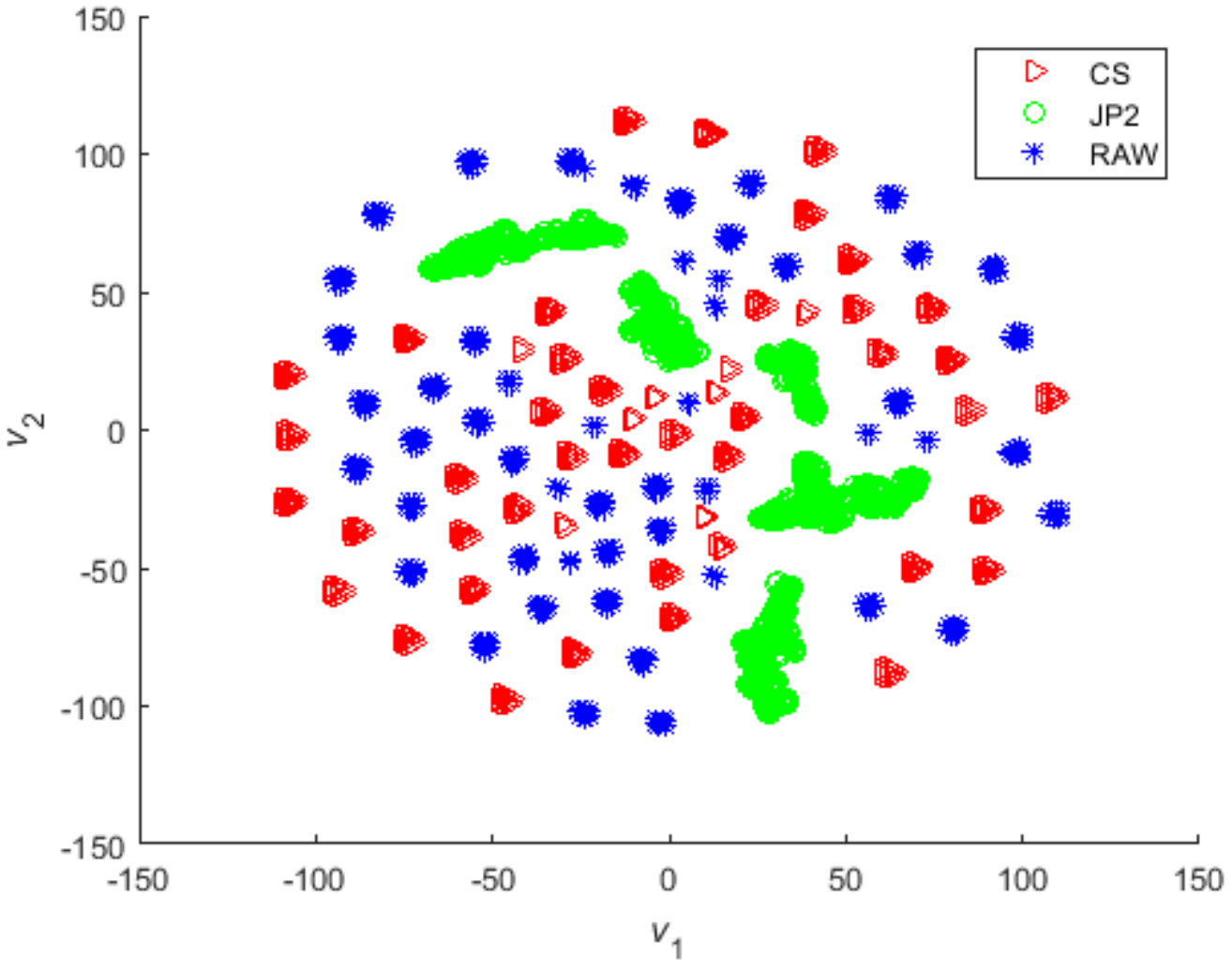}}
\hspace{0.5cm}%
%%----start of third subfigure----
\subfloat[]{
\label{fig:subfig:c} %% label for third subfigure
\includegraphics[width=5.55cm]{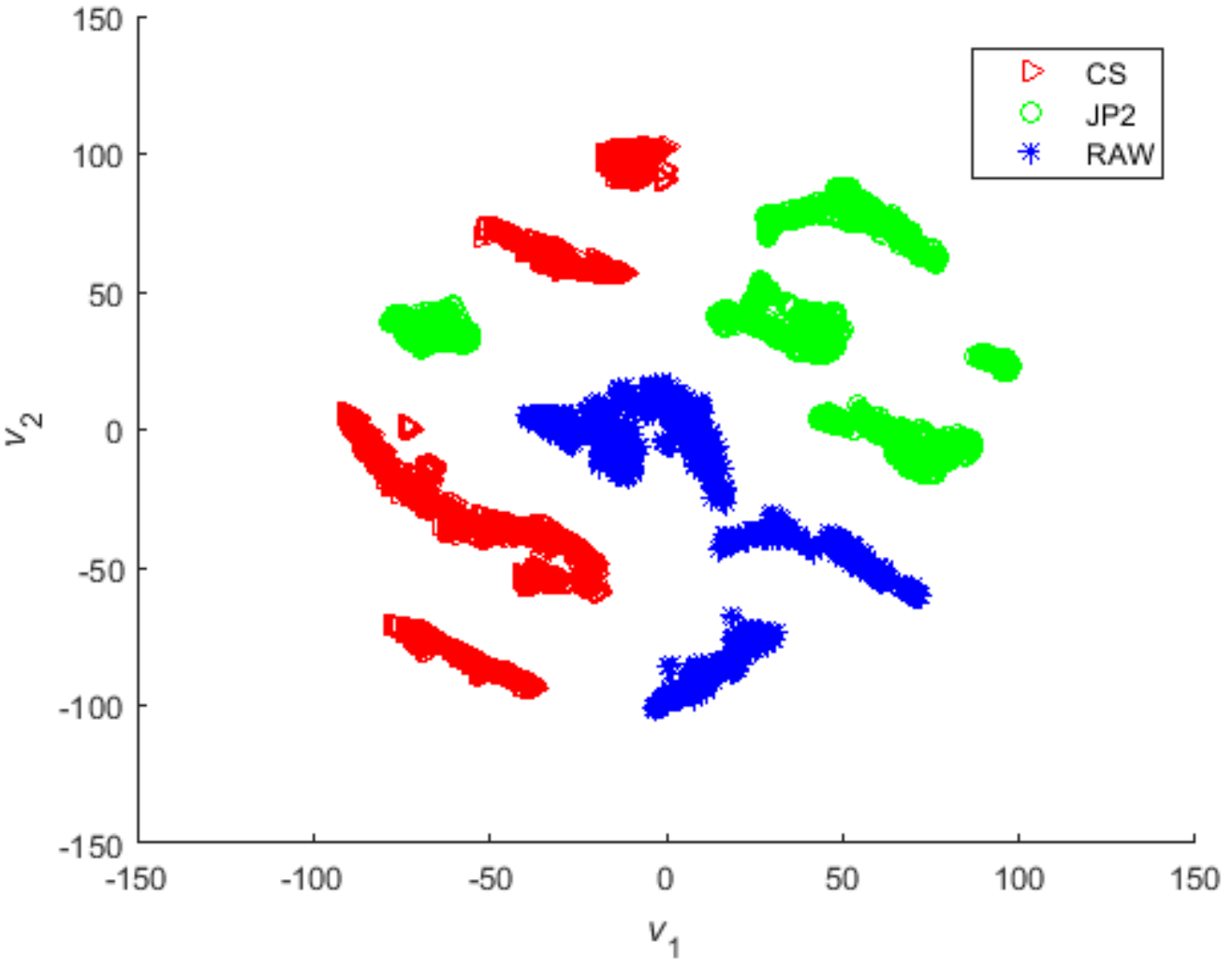}}
\caption{Scatterplots (a), (b), and (c) illustrate the 2-D visualization of blur kernel features for various imaging systems at 3 random runs.}
\label{VisualizationResults} %% label for entire figure
\end{figure*}

Here, we investigate the discriminability of the proposed signatures in a low-dimensional cluster representation by using the visualization tool presented in~\cite{taimori2016novel}. This visualizer is a bi-level dimensionality reduction technique. At the first level, the kernel features are sparsely coded via compressed sensing technology and group Least Absolute Shrinkage and Selection Operator (LASSO) algorithm. Using the unsupervised non-parametric t-SNE dimensionality reduction algorithm \cite{van2008visualizing}, sparse vectors are mapped onto a 2-D space with components $v_1$ and $v_2$ at the second level of the visualizer.

For visualization, we randomly selected with replacement $50\%$ of the database $\mathcal{D}$ for training and remainder for evaluation. In the visualizer, we adjusted the parameters of the number of measurements as $d=10$ and the regularizer as $\lambda=0.5$. Figure~\ref{VisualizationResults} illustrates the 2-D visualized results at 3 random runs. It is noticeable that due to stochastic nature of the visualization algorithm~\cite{taimori2016novel}, clustering results at various random runs are different. Visualized information reveal that
\begin{list}{\labelitemi}{\leftmargin=0.5em}
\item point clouds pertaining to class JP2 have separated well,
\item  data points of classes CS and RAW have entangled with each other, and
\item distinct islands convey species information belonging to individual classes. These species may be generated by different ratios of coders and other latent variables of imaging systems. We have analyzed the behavior of sub-classes in Section \ref{IdentificationQualityforVariousSettingsofCoders}.
\end{list}

\subsection{Identification Quality for Various Settings of Coders}
\label{IdentificationQualityforVariousSettingsofCoders}
It is important to obtain the performance of classification for sub-classes of each sensing system. The compressive ratio in compressed imaging and the compression ratio in JPEG2000 coder are main parameters for controlling the required bandwidth. Settings of these ratios directly affect the quality of images in their decoding phase. Hence, the estimated kernels vary based on the amount of compressive/compression ratios.

Figures~\ref{PerEvaluations} (b) and (c) compare the accuracy of different methods for detecting CS and JP2 classes under various compressive and compression ratios, respectively. In Fig.~\ref{PerEvaluations} (b), by increasing the compressive ratio, the accuracy for detecting class CS decreases. It is due to the fact that the higher the compressive ratio, the lower compressed sensing artifacts. This leads to increasing the error probability of misclassification class CS as class RAW and vice versa (See Table~\ref{ConfusionTable} for more information). Inversely, the performance of detecting class JP2 decreases by reducing the compression ratio. The lower the compression ratio, the lower compression artifacts, which increases the error probability of misclassification between classes JP2 and RAW. The value in parentheses in the legend of Figs.~\ref{PerEvaluations} (b) and (c) shows the average accuracy on all ratios. The our proposed kernel+deep CNN has the first accuracy rank among different methods. The competing method~\cite{chu2015compressive} exhibits almost the same accuracy for various ratios. However, its performance is lower than the suggested methods, especially for detecting class CS in Fig.~\ref{PerEvaluations} (b). The overall performance for detecting class JP2 is better that class CS, as was justified by visualization, too.

We implemented the method~\cite{chu2015compressive} and estimated the thresholding parameters $\tau_1$ and $\tau_2$ of the first and the second detectors from the learning set $\mathcal{D}_L$. In a grid search, we varied $\tau_1$ from $0.001$ to $0.002$ and $\tau_2$ from $0.002$ to $0.003$, and found the optimal $\tau_1^{*}=0.0015$ and $\tau_2^{*}=0.002$ values so that the accuracy constraint on learning data is maximized. In this approach, we also utilized 5 levels of wavelet decomposition which gives the best performance and set the number of iterations in EM algorithm equal to $100$.

\subsection{Detailed and Overall Performance Measures}
In this experiment, we compare the performance of modern approaches. We used confusion table to determine the details of probabilities of correct classification and misclassification. Table~\ref{ConfusionTable} reports the confusion tables of competing methods. In the case of 3-class classification problem, the classes CS, JP2, and RAW have denoted by $C_1$, $C_2$, and $C_3$, respectively. In the table, GT and PR acronyms are for Ground Truth and Predicted Result, respectively. Overall accuracy in terms of percent for kernel+deep CNN, kernel+SVM, image pixels+deep CNN, and Chu et al.~\cite{chu2015compressive} approaches is $80.28$, $75.51$, $82.01$, and $55.64$, respectively. This means that the methods of image pixels+deep CNN and kernel+deep CNN receive the first and the second ranks of accuracy, respectively.

Confusion tables show that for all methods, the misclassification errors between classes CS and RAW is more than other errors. The correct classification of class JP2 is also more than other systems, because blur kernels of JPEG2000 images are more discriminative. Therefore, the best accuracy has obtained for predicting the class JP2 and the worst accuracy for the class RAW. The accuracy of the kernel+deep CNN for detecting classes CS and JP2 is higher than the image+deep CNN, while for class RAW the problem is inverse. Thanks to the performance difference of class $C_3$, the overall accuracy of the image+deep CNN method is more than the kernel+deep CNN. This also shows that specifically for class RAW, the automatically extracted features by means of convolution plus pooling represent better discriminability than the blur kernels estimated from deconvolution.

\begin{table*}[!t]
\renewcommand{\arraystretch}{2.2}
\centering
\caption{Confusion tables for compressive identification of different approaches in $\%$}
\centering
\label{ConfusionTable}
\centering
\resizebox{18cm}{!}{
\begin{threeparttable}
\begin{tabular}{c c||c c c c|c c c c|c c c c}
\hline
\hline
\multicolumn{2}{c||}{\multirow{3}{*}{Confusion table}} & \multicolumn{12}{c}{$\text{GT}^{\tnote{\dag}}$} \\ \cline{3-14}
&  & \multicolumn{4}{c|}{$C_1$} & \multicolumn{4}{c|}{$C_2$} & \multicolumn{4}{c}{$C_3$} \\ \cline{3-14}
&  & Kernel+CNN	& Kernel+SVM	& Image+CNN	& Method~\cite{chu2015compressive}	 & Kernel+CNN	& Kernel+SVM	& Image+CNN	& Method~\cite{chu2015compressive}	& Kernel+CNN	& Kernel+SVM	& Image+CNN	& Method~\cite{chu2015compressive}\\
\hline
\hline
\multicolumn{1}{c|}{\multirow{3}{*}{$\text{PR}^{\tnote{\ddag}}$}}
& \multicolumn{1}{|c||}{$C_1$} & 83.84	& 78.96	& 79.42	& 54.88	& 1.07	& 1.68	& 1.68	& 2.13	& 32.01	& 34.45	& 12.96	& 46.49 \\
\cline{2-14}
& \multicolumn{1}{|c||}{$C_2$} & 0.24	& 0.61	& 3.05	& 19.66	& 94.05	& 89.94	& 87.04	& 81.86	& 5.03	& 7.93	& 7.47	& 23.32 \\
\cline{2-14}
& \multicolumn{1}{|c||}{$C_3$} & 16 	& 20.43	& 17.53	& 25.46	& 4.88	& 8.38	& 11.28	& 16.01	& 62.96	& 57.62	& 79.57	& 30.18 \\
\hline
\hline
\end{tabular}
\begin{tablenotes}
\item[\dag, \ddag] GT and PR are acronyms for Ground Truth and Predicted Result, respectively. In the table, the parameters $C_1$, $C_2$, and $C_3$ also denote classes CS, JP2, and RAW, respectively.
\end{tablenotes}
\end{threeparttable}}
\end{table*}

\section{Conclusions and Future Researches}
\label{ConclusionsAndFutureResearches}
This paper presents a framework for forensic identification of a compressive imaging system as well as discriminating it from traditional imagers. At first, we systematically modeled compressive and conventional imaging systems. Then, the abilities of footprints for discrimination between compressive imaging and conventional one were mathematically demonstrated. Theoretical analyses are supported by numerical simulations, too. Through the paper, we modeled the process of an image acquisition system as a deconvolution problem. Under the scenario that whether sensing has been done by means of a compressive imaging system or a tradition one, we shown an estimated blur kernel of a given imaging system provides useful information for identifying the history of a suspect image.

In order to improve the performance of the proposed identification method by learning from blur kernel signatures, complementary information such as noise characteristics and sharpness property of images can be unified the features. Our modeling also facilitates developmental approaches for investigation of forensic scenarios arose in the next-generation state-of-the-art compressive imaging systems.

\section*{Acknowledgements}
This Postdoc research was jointly sponsored by Iran National Science Foundation (INSF) and ACRI of Sharif University of Technology under agreement numbers 95/SAD/47585 and 7000/6642, respectively. We also thank Prof A. Amini and other researchers in Signal Processing and Multimedia Lab for their valuable comments.

\ifCLASSOPTIONcaptionsoff
  \newpage
\fi



%\bibliographystyle{IEEEtran}
%\begin{thebibliography}{1}
%\end{thebibliography}

\bibliographystyle{IEEEtran}
%\bibliographystyle{Ref}
% \bibliographystyle{spmpsci}
%\bibliography{IEEEabrv,../bib/Ref}

{\footnotesize
\bibliography{bare_jrnl}}

\end{document}